\documentclass[11pt]{article}
\usepackage[T1]{fontenc}
\usepackage[hmargin=0.85in,vmargin=0.80in]{geometry}
\usepackage{amsmath,booktabs,float,graphicx}
\usepackage[hidelinks]{hyperref}
\usepackage{microtype}
\newcommand{\rankingfoldrows}{%
1 & 164 & 54 & 2.036 & -0.004 & 1.846 & +0.011 \\
2 & 172 & 56 & 2.153 & +0.014 & 2.270 & -0.034 \\
3 & 153 & 52 & 3.345 & -0.017 & 3.343 & +0.043 \\
4 & 151 & 50 & 4.750 & +0.000 & 4.614 & -0.162 \\
5 & 171 & 53 & 2.690 & -0.048 & 3.208 & -0.082 \\
}
\newcommand{\frozenfoldrows}{%
1 & 166 & 156/164 & 156/164 & 151/164 & 152/164 \\
2 & 156 & 157/172 & 156/172 & 157/172 & 163/172 \\
3 & 167 & 136/153 & 137/153 & 138/153 & 135/153 \\
4 & 165 & 136/151 & 137/151 & 141/151 & 139/151 \\
5 & 171 & 150/171 & 148/171 & 161/171 & 158/171 \\
}

\newcommand{\studyrepo}{\href{https://github.com/naazzarov/BUS-BRA-Granular-Computing-Personal/tree/research/patient-deferral/research/patient_deferral}{\nolinkurl{naazzarov/BUS-BRA-Granular-Computing-Personal}}}
\title{Does Image Disagreement Help Decide Which Breast Ultrasound Predictions to Defer?}
\author{Bahodir Nazarov, Zineb Gharib, Fatima Zazoul, and Luo Chuan}
\hypersetup{
  pdftitle={Does Image Disagreement Help Decide Which Breast Ultrasound Predictions to Defer?},
  pdfauthor={Bahodir Nazarov, Zineb Gharib, Fatima Zazoul, and Luo Chuan}
}
\date{}
\begin{document}
\maketitle

\begin{abstract}
\textbf{Purpose.} We test whether disagreement between two breast ultrasound image scores helps rank patient-level errors after accounting for confidence in the average score.

\textbf{Methods.} We use 1,875 image records from 1,064 BUS-BRA patient groups, including 811 two-image groups. We compare confidence-only and confidence-plus-disagreement error rankers using patient-disjoint nested splits with logistic regression and histogram gradient boosting. Secondary analyses test frozen cutoffs, held-out scanners, and an alternate feature set generated from the public archive.

\textbf{Results.} Adding disagreement changed area under the generalized risk--coverage curve by $-0.011$ percentage points with logistic regression (paired patient-bootstrap 95\% interval $[-0.056,0.033]$) and by $-0.045$ points with gradient boosting ($[-0.131,0.038]$). With a confidence-only cutoff set to retain at least 90\% of separate calibration patients, the logistic model retained 90.6\% of test patients overall, including 83.4\% of malignant and 94.1\% of benign cases. The benign-minus-malignant coverage gap was 10.7 points (95\% interval $[6.1,15.5]$).

\textbf{Conclusion.} The incremental benefit of absolute score disagreement was uncertain in this cohort. The frozen cutoff retained fewer malignant cases proportionally, supporting diagnosis-specific reporting alongside overall coverage.

\textbf{Code.} Analysis code and aggregate results are maintained in the private GitHub repository \studyrepo.
\end{abstract}

\section{Introduction}
A selective classifier can retain or defer a prediction. As it defers more cases, its errors among retained cases can fall while the number of cases referred for review rises \cite{geifman}. In breast ultrasound, a patient may have multiple images of a finding. Two image scores can agree, straddle the classification boundary, or diverge while supporting the same decision. These patterns suggest that disagreement might help identify errors even when the average score is available.

Confidence and disagreement can encode overlapping information. For example, two scores whose average is close to one cannot be far apart. The practical question is whether the available disagreement adds enough information to change which errors a model retains. We test that question using patient-level decisions held fixed within each comparison. The resulting comparison isolates the choice of cases to defer from changes to the classifier itself.

A second question concerns the review queue. A rule that retains a target fraction of patients can defer different fractions of benign and malignant findings. We estimate diagnosis-specific coverage and report errors remaining among retained cases. The test uses the BUS-BRA cohort \cite{busbra}, with a second protocol that fixes the cutoff using separate calibration patients. Scanner holdouts and an alternative input representation test sensitivity within this cohort.

\section{Related work}
Wei et al.\ combined two ultrasound views to classify breast findings \cite{wei}. That study evaluated differences in the decisions produced by a classifier. Our analysis assigns a fixed decision to each patient, then asks whether the two image scores improve ranking of those decisions by error risk.

Geifman and El-Yaniv formalized the selective-classification trade-off between risk and coverage \cite{geifman}. Traub et al.\ proposed AUGRC to summarize generalized risk over the coverage range \cite{traub}. Jones et al.\ demonstrated that selective classification can affect groups unevenly \cite{jones}. We use pathology labels to audit review-queue composition. Pathology is observed retrospectively and enters the evaluation of coverage by diagnosis.

Kinalioglu evaluated conformal reliability on held-out BUS-BRA scanners \cite{kinalioglu}. Our device analysis examines patient-level error ranking and the coverage reached by a cutoff fitted on other devices. Conformal risk control offers a framework for specifying a formal risk target under stated assumptions \cite{angelopoulos}. The present replay measures achieved coverage for an empirical cutoff.

\section{Materials and methods}
\subsection{Data and training boundaries}
BUS-BRA contains 1,875 anonymized B-mode images from 1,064 patients at the Brazilian National Cancer Institute. Its benign or malignant reference diagnosis was established by biopsy \cite{busbra}. The derived table contains 102 original-image features and 186 features from first-level sym2 wavelet images. These were extracted within the supplied lesion masks using 2D PyRadiomics with intensity bin width 16 and no normalization or spatial resampling. Shape measurements use pixel units because physical pixel spacing is unavailable. We verified the table's image, patient, diagnosis, and five-fold assignments against the official release \cite{zenodo}.

There are 253 patient groups with one image and 811 with two images. Each group has one diagnosis and one outer fold. The two-image cohort contains 265 malignant and 546 benign groups. The primary analysis uses these 811 groups because disagreement requires two images. Single-image groups also contribute to training. The training unit for the classifier is an image and the sampling unit for all splits and uncertainty estimates is a patient group. We use \emph{two-image} to describe dataset records belonging to the same patient, regardless of acquisition plane. Patient and image keys, scanner, and fold assignments are excluded from predictors.

\begin{figure}[ht]
\centering
\includegraphics[width=\linewidth]{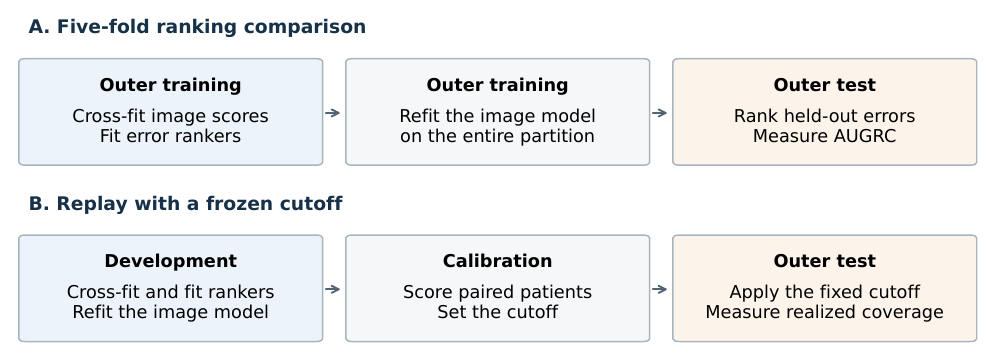}
\caption{Patient partitions for ranking and cutoff replay. In A, both ranking rules score the same fixed test decisions. In B, model fitting and cutoff selection use separate patient groups. A held-out scanner replaces the outer fold in the device analysis.}
\label{fig:partitions}
\end{figure}

We retain the five predefined patient-disjoint outer folds (Figure~\ref{fig:partitions}A). For each fold, a shuffled three-fold stratified group split within its training data produces an out-of-fold image probability for every outer-training image. The random seed for the inner split is 42 plus the outer fold number. Each of the three inner models scores patients it did not train on. As a result, the error labels for ranker fitting reflect predictions made on training patients held out from the relevant image model.

A patient contributes total weight one to image model fitting, divided equally over the patient's images. The primary classifier is L2-regularized logistic regression with $C=1$ and training-weighted standard scaling. The fixed histogram gradient boosting model uses 150 iterations, at most 15 leaves per tree, L2 regularization 1, and seed 42. For each outer fold, after fitting the rankers on the inner predictions, we refit the image model on all outer-training records and generate one set of patient predictions for the held-out fold. The prediction and weighting rules are common to both models.

\subsection{Patient decision and deferral rules}
For image scores $p_1$ and $p_2$, the patient score is $\bar p=(p_1+p_2)/2$. A score of at least 0.5 assigns a malignant prediction. The confidence margin is $m=|\bar p-0.5|$ and the disagreement is $d=|p_1-p_2|$. The two quantities obey
\[
d \le 2\min(\bar p,1-\bar p)=1-2m.
\]
To see the bound, order the scores as $p_{\mathrm{lo}} \le p_{\mathrm{hi}}$. Their gap is at most $p_{\mathrm{hi}}+p_{\mathrm{lo}}$ and at most $(1-p_{\mathrm{hi}})+(1-p_{\mathrm{lo}})$. Consequently, large absolute disagreement is feasible only over a restricted range of average confidence. The confidence-only comparison accounts for the average margin when asking whether the observed position within that range predicts errors.

We use two standardized logistic error models with $C=1$, fitted to errors in the inner out-of-fold predictions of two-image training patients. One uses $m$ alone and the other uses $m$ and $d$. Each estimates the probability that the fixed patient decision is incorrect. Low estimated error risk ranks a patient earlier for acceptance. We also report two rules that require no second model. They rank by higher $m$ or by the higher minimum image margin $\min_j|p_j-0.5|$. All rules use the same underlying patient predictions within each backbone and outer fold.

Each learned ranker is fitted to one row per paired patient. The target is the indicator that the pair-average decision differs from the reference diagnosis. The standardized margin and disagreement enter as additive terms on the log-odds scale. The analysis tests that specified representation of disagreement. It leaves open interactions, asymmetric image weighting, and rankers trained on explicit image quality.

\subsection{Outcomes and uncertainty}
For $n$ test patients and a threshold $t$ on estimated error risk, let $a_i(t)$ indicate retention and let $e_i$ indicate a classification error. Define coverage and generalized risk by
\begin{equation}
C(t)=\frac{1}{n}\sum_{i=1}^{n} a_i(t), \qquad
G(t)=\frac{1}{n}\sum_{i=1}^{n} a_i(t)e_i.
\end{equation}
The error rate conditional on retention is $G(t)/C(t)$ for $C(t)>0$. The primary endpoint integrates generalized risk over the coverage range \cite{traub}. If $r_i$ is the rank of patient $i$ in order of increasing estimated error risk, average ranks for ties give
\begin{equation}
\mathrm{AUGRC}=\frac{1}{n^2}\sum_{i=1}^{n} e_i\left(n-r_i+\tfrac{1}{2}\right).
\end{equation}
The rank formula is the trapezoidal integral of the empirical generalized risk. A smaller value moves errors toward the end of the ranking. We calculate one value per outer fold and average with equal fold weight. This avoids sorting scores from different fitted rankers on a single shared scale.

At nominal 80\% and 90\% coverage, we round each fold's retained count up and average counts over a uniform tie break at the boundary. This analysis sets the retained count using the test fold size. For context, a uniformly random selection of $k$ from $n$ patients with $E$ errors would retain $kE/n$ errors on average. This descriptive calculation uses the same fold-specific retention quotas. Under the frozen-cutoff protocol, we additionally report the fraction retained in each diagnosis, the number of errors among retained cases, and how many malignant cases remain classified as benign.

The primary contrast is AUGRC for the confidence-plus-disagreement rule minus AUGRC for the confidence-only rule with the logistic backbone. We draw 5,000 paired bootstrap samples of patient groups within outer folds using seed 20261124. Each draw uses the same resampled patients for both rules. The 2.5th and 97.5th percentiles give a conditional interval for the fold-mean difference. The bootstrap holds the predictions and fitted models fixed \cite{rutter}. We recorded the evaluation plan before inspecting this study's outcomes. The plan was kept internally.

\subsection{Secondary analyses}
The primary results were available when we designed the secondary analyses. For each official outer fold, we reserved one block of a four-way grouped split of the outer training data for cutoff calibration (Figure~\ref{fig:partitions}B). We fitted the image model and error rankers in the remaining development data using the same three-way grouped cross-fitting for the rankers. For $n_{\mathrm{cal}}$ paired calibration patients, the cutoff was the estimated error score at rank $\lceil 0.90 n_{\mathrm{cal}}\rceil$ when sorted from low to high. This threshold used only estimated error scores. We included all ties at the cutoff and applied the unchanged models and cutoff to the outer test fold. Test coverage was an observed outcome. For the pooled benign-minus-malignant coverage gap, we computed a conditional percentile interval from 5,000 patient resamples within test folds (seed 20261125).

In the scanner analysis, we held out one acquisition device at a time and applied the same development and calibration procedure to patients scanned on the other devices. A test device needed at least 50 paired patients and at least 10 per diagnosis. Patients with ambiguous device assignment were ineligible for this analysis. The calibration and development cross-fitting seeds were 6040 and 7040 plus the outer fold number or the device's position in alphabetical order, respectively.

To test sensitivity to the feature representation, we reran the original ranking comparison with 12 descriptors computed directly from the public images and lesion masks. These comprised seven grayscale intensity summaries (mean, population standard deviation, and the 10th, 25th, 50th, 75th, and 90th percentiles), the mean absolute difference between horizontally or vertically adjacent foreground pixels, and four mask measurements (width, height, area, and bounding-box fill). Grayscale intensities were divided by 255. Bounding-box fill is the fraction of pixels inside the bounding box that belong to the mask. This replay uses the same patients and official folds.

\section{Results}
\subsection{Error ranking at matched coverage}
At full coverage the logistic backbone achieved 88.4\% accuracy and pooled area under the ROC curve 0.930 on two-image groups. Gradient boosting achieved 88.0\% accuracy and pooled ROC area 0.926. In single-image groups the observed accuracy was 84.2\% for both backbones.

\begin{table}[ht]
\centering
\small
\caption{Mean AUGRC across the five outer folds, in percent. Lower values indicate fewer retained errors over coverage levels.}
\label{tab:augrc}
\begin{tabular}{lrr}
\toprule
Ranking rule & Logistic & Gradient boosting\\
\midrule
Pair-average margin & 2.995 & 3.056\\
Minimum image margin & 3.141 & 3.390\\
Learned confidence only & 2.995 & 3.056\\
Learned confidence plus disagreement & 2.984 & 3.011\\
\bottomrule
\end{tabular}
\end{table}

\begin{figure}[ht]
\centering
\includegraphics[width=0.93\linewidth]{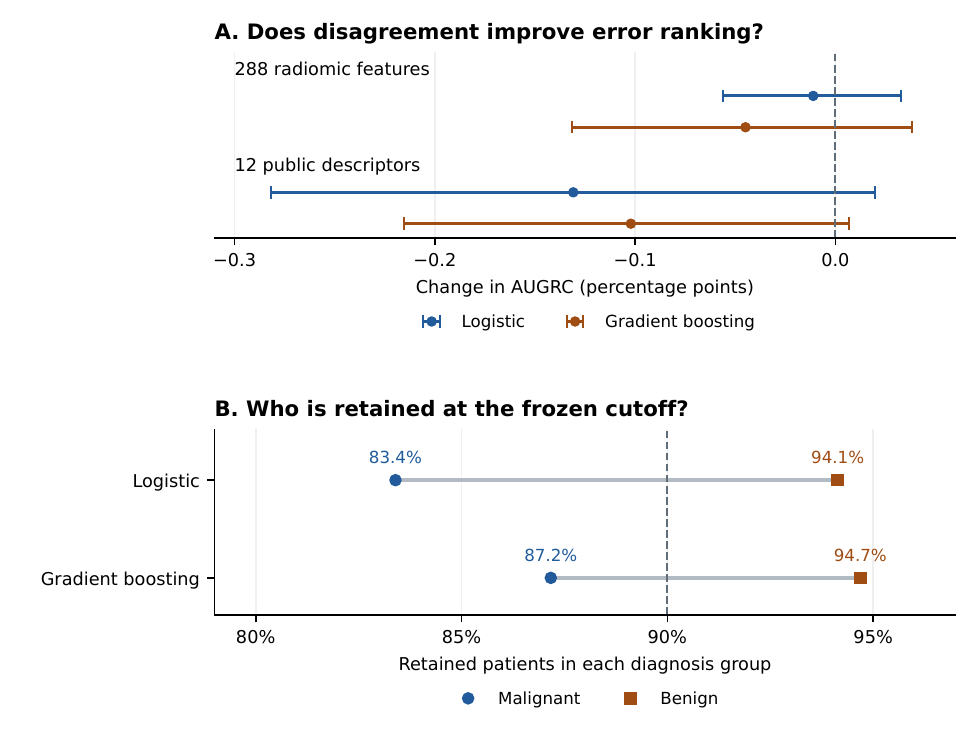}
\caption{A. Difference in fold-mean AUGRC for confidence plus disagreement versus confidence alone, with conditional paired 95\% intervals. Negative values favor the disagreement rule. B. Diagnosis-specific coverage for the confidence-only rule at cutoffs frozen on separate calibration patients. The dashed lines mark zero difference in A and the nominal coverage target in B.}
\label{fig:results}
\end{figure}

Adding disagreement changed AUGRC by $-0.011$ percentage points for the logistic backbone and by $-0.045$ for gradient boosting. The paired intervals span zero in both cases (Figure~\ref{fig:results}A). The confidence-only ranker and the untrained pair-average margin produced the same mean AUGRC for each model to three decimals. The absolute change when adding disagreement to the learned margin was below 0.05 points in every logistic fold. Confidence-only AUGRC varied more across folds, ranging from 2.036\% to 4.750\% for logistic regression (Table~\ref{tab:rankingfold}).

The logistic confidence-only and disagreement rules each retained 67 errors among 731 accepted patients at nominal 90\% coverage. At nominal 80\% coverage they retained 43 and 42 errors among 651 accepted patients, respectively. The expected error counts under uniform selection at those same fold quotas were 84.7 at 90\% and 75.4 at 80\%. The gradient boosting rules retained 65 and 64 errors among 731 accepted patients at nominal 90\% coverage, compared with 87.4 expected under uniform selection. Confidence concentrated errors in the deferred fraction. Adding disagreement changed few retained error counts at these operating points.

With confidence-only ranking at nominal 90\% coverage, logistic regression retained 219 of 265 malignant groups (82.6\%) and 512 of 546 benign groups (93.8\%). Gradient boosting retained 215 malignant groups (81.1\%) and 516 benign groups (94.5\%). Rounding within folds resulted in 90.1\% pooled coverage for both backbones.

\subsection{Frozen cutoffs and review-queue composition}
The frozen confidence-only cutoff retained 735 of 811 patients (90.6\%) with logistic regression. It retained 221 of 265 malignant cases (83.4\%) and 514 of 546 benign cases (94.1\%). The coverage gap was 10.7 percentage points (conditional 95\% interval $[6.1,15.5]$). With gradient boosting the gap was 7.5 points ($[3.2,12.3]$), with 748 of 811 patients retained (92.2\%). Fold-level coverage ranged from 87.7\% to 95.1\% for logistic regression and from 90.2\% to 94.2\% for gradient boosting (Table~\ref{tab:replayfold}).

The logistic confidence-only rule deferred 76 patients, including 44 malignant cases. Malignant findings made up 57.9\% of the deferred queue and 32.7\% of the paired cohort. The analogous fraction in the gradient boosting queue was 54.0\% (34 of 63). Among the 221 retained malignant cases in the logistic confidence-only replay, 46 (20.8\%) received a benign prediction. This was 17.4\% of all 265 malignant test cases when deferred patients remained in the denominator. Adding disagreement reduced the number of retained malignant cases by one for logistic regression and two for gradient boosting (Table~\ref{tab:frozen}).

\subsection{Transport to held-out scanners}

Every patient had a consistent device assignment. Three devices qualified for held-out testing, covering 800 paired patients. The fourth device (U-Systems) had 11 paired patients, of whom five were malignant. Across the three scanner holdouts, test coverage ranged from 92.5\% to 95.9\% and changes in AUGRC attributable to adding disagreement ranged from $-0.055$ to $+0.062$ points (Table~\ref{tab:devices}). The direction of the AUGRC change varied by device and backbone.

For example, the logistic disagreement rule ranked errors marginally better on GE Logiq 5 by AUGRC. At the cutoff learned on other scanners, it accepted nine more cases and retained three more errors. On GE Logiq 7 it accepted three fewer cases and retained the same number of errors. An integrated ranking metric and a fixed-cutoff error count can move in different directions because threshold transport also changes achieved coverage.

\subsection{Public-input sensitivity analysis}
With the 12 public descriptors, full-coverage accuracy was 84.8\% for logistic regression and 83.5\% for gradient boosting. The respective ROC areas were 0.908 and 0.888. The disagreement rule changed AUGRC by $-0.131$ points for logistic regression ($[-0.282,0.020]$) and $-0.102$ points for gradient boosting ($[-0.215,0.007]$). Both intervals include zero.

\section{Discussion}
\subsection{What disagreement adds}
The two backbones yielded small estimated changes in error ranking when we added absolute disagreement to confidence. The public descriptors yielded larger estimated improvements and wider intervals. Their lower full-coverage accuracy also left more errors available to rank. The confidence margin and disagreement share a geometric constraint. The results apply to absolute disagreements under the evaluated rankers and do not establish a practically negligible effect across models. In particular, the study leaves open whether a ranker with interactions, a direction-sensitive score, or independent image-quality information could improve decisions.

\subsection{Interpreting coverage and errors together}
The frozen cutoff retained a smaller fraction of malignant cases for both backbones. More malignant findings would therefore reach the review queue per deferred case than their share of this cohort suggests. Depending on the review process, this concentration might improve the yield of review or increase demand for timely review. The 46 retained false negatives in the logistic confidence-only replay also matter. Their number depends on the classification boundary and the cutoff jointly. Reporting retention without retained errors leaves out this part of the workflow.

Diagnosis determines the denominator for this audit. In practice it is unavailable at the time of deferral, and equal retention by diagnosis is not an objective specified by the study. Prediction quality after human review would also depend on which readers see the deferred cases, what additional imaging is available, and how disagreements are resolved. We measured retained model errors and queue composition under a retrospective replay. A workflow-level evaluation would require prospective or reader-study outcomes and a specified clinical decision. Decision-curve methods offer one way to relate model predictions to an explicitly defined decision and its relative costs \cite{vickers}.

\subsection{Scope of transfer and uncertainty}
The primary analysis uses two-image groups from one institution. The inputs rely on existing lesion masks and the results apply to the supplied images after lesion localization. We summarize single-image performance only at full coverage. The biopsy-confirmed cases and image construction in BUS-BRA also limit generalization to a screening population or other settings with a different referral mix.

The conditional bootstrap fixes the tested predictions and omits variation from model refitting on overlapping training samples. The scanner holdouts share one source institution. Device, acquisition procedure, and case mix are entangled in this design. Differences across devices describe observed transfer and cannot isolate a scanner effect. The fourth scanner contributes too few paired cases for the held-out analysis.

The frozen cutoff targets a share of calibration patients. Its realized test coverage depends on the score distribution in the target group. Extending this analysis calls for an independent cohort with linked patient images, reproducible descriptors, and reference diagnoses. External testing should preserve the fitted ranker and cutoff for its first analysis and report the realized overall and diagnosis-specific coverage with uncertainty. A subsequent recalibration study could evaluate the value of local data.

\section{Conclusion}
In 811 BUS-BRA patient groups with two images, we found uncertain incremental gains from adding absolute score disagreement to confidence. Diagnosis-specific retention revealed a difference that the overall coverage estimate would obscure, even when the cutoff was fixed using separate calibration patients.

\section{Data and code availability}
The source collection and its publication describe acquisition and release conditions \cite{busbra,zenodo}. This secondary computational study used previously released anonymized data. Exact replication of the radiomic analysis requires authorized access to the frozen derived feature table. The public-input sensitivity analysis can be reproduced from the official BUS-BRA archive with the alternative feature extractor and the same evaluation code.

The study code is maintained on GitHub in \studyrepo.
The repository contains the runners, input checksums, analysis plans, environment requirements, and aggregate outputs. Repository access is currently restricted. The fold-level appendix provides observed counts and estimates for readers evaluating the analysis from the preprint. This reporting follows the distinction between internal and external testing in the CLAIM guidance \cite{claim}.

OpenAI Codex assisted implementation, testing, and manuscript preparation.

\appendix
\section{Fold-level results}
The patient counts below refer to two-image groups. In the ranking analysis, both rules use the same patient-level predictions within a backbone and fold. In the cutoff replay, the calibration counts are paired patients scored by the corresponding development model. The fractions in the test columns use every paired patient in the official test fold as the denominator.

\begin{table}[H]
\centering
\small
\caption{Per-fold generalized risk--coverage results, expressed as percentages and percentage-point changes. Base is the confidence-only AUGRC. A negative change favors adding disagreement.}
\label{tab:rankingfold}
\begin{tabular}{rrrrrrr}
\toprule
& & & \multicolumn{2}{c}{Logistic} & \multicolumn{2}{c}{Gradient boosting}\\
\cmidrule(lr){4-5}\cmidrule(lr){6-7}
Fold & Paired $n$ & Malignant $n$ & Base & $\Delta$AUGRC & Base & $\Delta$AUGRC\\
\midrule
\rankingfoldrows
\bottomrule
\end{tabular}
\end{table}

\begin{table}[H]
\centering
\small
\caption{Frozen cutoff replay by outer fold. Each cell shows the number of paired test patients retained and the number eligible. Base uses confidence alone. With gap adds absolute image-score disagreement.}
\label{tab:replayfold}
\begin{tabular}{rrrrrr}
\toprule
& & \multicolumn{2}{c}{Logistic} & \multicolumn{2}{c}{Gradient boosting}\\
\cmidrule(lr){3-4}\cmidrule(lr){5-6}
Fold & Calibration pairs & Base & With gap & Base & With gap\\
\midrule
\frozenfoldrows
\bottomrule
\end{tabular}
\end{table}

\section{Operating points for the follow-up analyses}

\begin{table}[H]
\centering
\small
\caption{Frozen cutoff replay. Counts describe 811 paired test patients. Coverage is the percentage retained within each diagnosis. FN denotes retained malignant cases classified as benign.}
\label{tab:frozen}
\begin{tabular}{llrrrrr}
\toprule
Backbone & Rule & Retained & Malignant & Benign & Errors & FN\\
\midrule
Logistic & Confidence & 735 & 83.4\% & 94.1\% & 64 & 46\\
Logistic & With disagreement & 734 & 83.0\% & 94.1\% & 65 & 46\\
Boosting & Confidence & 748 & 87.2\% & 94.7\% & 70 & 53\\
Boosting & With disagreement & 747 & 86.4\% & 94.9\% & 69 & 50\\
\bottomrule
\end{tabular}
\end{table}

\begin{table}[H]
\centering
\small
\caption{Held-out scanner performance. The first number in each pair is for confidence alone and the second adds disagreement. AUGRC differences are in percentage points.}
\label{tab:devices}
\begin{tabular}{llrrrr}
\toprule
Test scanner & Backbone & Paired $n$ & Coverage (\%) & Retained errors & $\Delta$AUGRC\\
\midrule
GE Logiq 5 & Logistic & 319 & 93.1 / 95.9 & 31 / 34 & $-0.036$\\
GE Logiq 5 & Boosting & 319 & 95.3 / 94.0 & 23 / 22 & $+0.044$\\
GE Logiq 7 & Logistic & 414 & 93.2 / 92.5 & 68 / 68 & $+0.062$\\
GE Logiq 7 & Boosting & 414 & 94.0 / 94.4 & 46 / 46 & $-0.055$\\
Toshiba Aplio 300 & Logistic & 67 & 92.5 / 94.0 & 8 / 9 & $+0.045$\\
Toshiba Aplio 300 & Boosting & 67 & 92.5 / 95.5 & 10 / 11 & $+0.022$\\
\bottomrule
\end{tabular}
\end{table}

\end{document}